\pdfoutput=1
\documentclass[runningheads]{llncs}

\usepackage{accv}
\usepackage{accvabbrv}
\usepackage{graphicx}
\usepackage{booktabs}
\usepackage{hyperref}
\usepackage{orcidlink}
\usepackage[capitalize]{cleveref}

\newcommand{\ci}[2]{{\footnotesize[#1,\,#2]}}
\newcommand{\cis}[2]{{\scriptsize[#1,\,#2]}}

\usepackage{eso-pic}
\AddToShipoutPicture*{%
  \AtPageLowerLeft{%
    \put(\LenToUnit{0.5\paperwidth},\LenToUnit{0.5in}){%
      \makebox[0pt][c]{\parbox[b]{\textwidth}{\centering\scriptsize
        Accepted to the ACCV 2026 Workshop on Subspaces in the Era of
        Foundation Models.\\
        This version of the contribution has been accepted for publication,
        after peer review, but is not the Version of Record and does not
        reflect post-acceptance improvements, or any corrections. The Version
        of Record will appear in the ACCV 2026 Workshops proceedings
        (Springer LNCS).}}}}}

\begin{document}

\title{Subspace Representations for Synthetic-to-Real Transfer
in 3-D Surface Inspection}
\titlerunning{Subspace Representations for Synthetic-to-Real Transfer}

\author{Abdulla Nasir Chowdhury\inst{1} \and Naish Ibrahim\inst{1} \and
Gulpi Qorik Oktagalu Pratamasunu\inst{1} \and Kazuhiro Fukui\inst{2}}
\authorrunning{A.~N.~Chowdhury et al.}
\institute{Department of Computer Science, University of Tsukuba, Japan\\
\email{\{abdullah.nasir,naish,gulpi\}@cvlab.cs.tsukuba.ac.jp}
\and
Tsukuba Institute for Advanced Research, Department of Computer Science,
University of Tsukuba, Japan\\
\email{kfukui@cs.tsukuba.ac.jp}}
\maketitle

\begin{abstract}
\emergencystretch=1em
This paper studies what subspace representations contribute to
synthetic-to-real transfer in 3-D surface inspection, where models
learn from synthetic defects and the only real data before deployment
are scans of good parts. We use three classical constructions as
instruments: the nearest-subspace classifier, one-class template
subspaces and the generalized difference subspace (GDS). On 1{,}197
MVTec 3D-AD scans without target annotation we find that every
decision rule ranks real defects at a similar window AUC, that
image-level detection stays at chance, and that no rule keeps its
in-domain threshold. We show that the margin of a nearest-subspace rule
moves by at most the part of the feature shift outside its class
subspaces, and that this leakage, measured from unlabeled good scans,
bounds the displacement observed on unseen scans. On 597 Real3D-AD objects,
a one-class subspace of registered templates evaluated at every
location on frozen features reaches point AUROC 0.957 on pre-registered
scans and a category-mean object AUROC of 78 to 81 with no training on
the benchmark. Our results suggest that a subspace helps under shift in
proportion to the shift it spans, and that in
reference-based inspection the reference and its registration matter
more than the subspace.
\end{abstract}

\section{Introduction}
\label{sec:intro}
In this paper, we ask what subspace representations contribute to
synthetic-to-real transfer. We study the question in 3-D surface
inspection, where models are trained on synthetic defects because real
defects are rare and costly to annotate, and where the only real data
available before deployment are scans of good parts. DentNet
\cite{lafiosca2022}, for example, segments dents on \emph{residual
images}, the deviation from a fitted quadric, and reaches near-perfect
IoU on its synthetic benchmark, but its behavior on real scans of other
objects has not been measured.

A subspace suppresses the variation inside its span and leaks the
rest, so what it contributes to the transfer depends on where the
synthetic-to-real variation lies. A subspace can enter the transfer in
two places. It can classify the features of an encoder trained on
synthetic data, and then the variation it must absorb is the feature
shift between the synthetic and the real domain. It can also model the
real good parts directly, and then the variation is the disagreement
among good templates and the error of registering a scan to them. We
study both cases with the same classical constructions, the
nearest-subspace classifier, a one-class subspace of good templates and
the generalized difference subspace (GDS) \cite{fukui2023generalized},
used as instruments rather than as new methods.

We hold the synthetic training fixed and vary the encoder and the
decision rule. Three learned families are compared: DentNet
\cite{lafiosca2022}, an end-to-end U-Net on residual images,
PointNet\texttt{++} \cite{qi2017pointnetpp} trained from scratch on the
identical windows as point sets, and patch-based subspace segmentation
(PBSS), which classifies DentNet's frozen features by nearest PCA
subspace, with frozen DINOv2 \cite{oquab2024dinov2} features as a
foundation-model representation and a training-free surface-variation statistic
\cite{pauly2002curvature} as the geometric baseline. Two public
benchmarks supply the real scans. MVTec 3D-AD \cite{bergmann2022mvtec3d}
provides no registered reference and tests the subspace as a
classifier, under a protocol that uses no target annotation. Real3D-AD
\cite{liu2023real3dad} provides registered good templates and tests the
subspace as a model of the good parts. In both, we measure the
variation itself and where it lies relative to the subspace, and we
separate what a ranking metric shows from what a fixed threshold shows,
since the two answer the question differently.

Our main contributions are summarized as follows:
\begin{itemize}
  \item As a classifier on synthetic-trained features, a subspace does
  not rank real defects better than other rules, but it makes the
  threshold less sensitive to the part of the shift it spans. We show
  that its margin moves by at most the leakage of the shift outside its
  class subspaces, a quantity measurable from unlabeled good scans, and
  that on MVTec 3D-AD this leakage bounds and ranks the displacement
  observed on unseen scans of every class.
  \item As a model of the real good parts, a one-class subspace of
  registered templates turns a frozen synthetic-trained encoder into a
  detector of real defects on Real3D-AD, where the classifier fails,
  with no training on the benchmark. We attribute the result step by
  step and find that the reference and its registration contribute more
  than the choice of subspace construction.
  \item We show that a fixed threshold, the window scale and the object
  aggregation rule can each invert a method ordering, and we report
  every result under controls that make the ordering visible.
\end{itemize}

The rest of this paper is organized as follows. \Cref{sec:related}
reviews related work. \Cref{sec:methods} describes the data, the
representations, and the subspace as a classifier and as a model of
the good parts, and states the leakage bound.
\Cref{sec:protocol} gives the evaluation protocol. \Cref{sec:results}
reports the in-domain results, the subspace as a classifier on MVTec
3D-AD and Real3D-AD, and the subspace as a model of the good parts on
Real3D-AD. \Cref{sec:discussion} answers the question and discusses
its limitations, and \cref{sec:conclusion} concludes the paper.

\section{Related Work}
\label{sec:related}
\textbf{Subspace methods.} Classification by subspace distances dates to
CLAFIC \cite{watanabe1973subspace,oja1983subspace} and Eigenfaces
\cite{turk1991eigenfaces}. The mutual subspace method compares two
subspaces by their canonical angles \cite{yamaguchi1998msm,fukui2007komsm},
and the generalized difference subspace \cite{fukui2023generalized}
isolates the directions on which a set of subspaces differ.
Subspace classifiers over deep features appear in few-shot recognition
\cite{simon2020adaptive}, and a PCA subspace of frozen DINOv2
\cite{oquab2024dinov2} patch features scored by reconstruction residual
is a strong training-free few-shot anomaly detector on images
\cite{lendering2026subspacead}. Both places where we use a subspace come
from this line of work: the classifier of \cref{sec:classifier} is the subspace
method itself, and the nominal subspace of \cref{sec:nominalmethod} is its
one-class form placed at a location.

\textbf{Synthetic-to-real transfer.} Domain randomization
\cite{tobin2017domain} and adversarial adaptation \cite{ganin2016dann}
enlarge or align training distributions. Subspace alignment
\cite{fernando2013subspace}, the geodesic flow kernel
\cite{gong2012geodesic} and CORAL \cite{sun2016coral} rest on the shift
between domains being carried by leading principal directions. We
measure that observation from the classifier's side with training fixed,
and the leakage of \cref{eq:leak} is its quantitative form for a
nearest-subspace rule.

\textbf{3-D surface inspection and anomaly detection.} Hand-crafted curvature features
\cite{jovancevic2017defects} preceded DentNet \cite{lafiosca2022}. Point-based
architectures \cite{qi2017pointnetpp,thomas2019kpconv} consume raw point
sets. MVTec 3D-AD \cite{bergmann2022mvtec3d} and Real3D-AD
\cite{liu2023real3dad} provide real scans with per-pixel masks and
per-point labels. The Real3D-AD baseline
registers each test scan to a nominal template and scores distances to
feature memory banks. The later methods that lead its table either
register as well \cite{zhu2024group3ad,yu2025reg2inv} or learn from
synthetic anomalies \cite{li2024imrnet,zhou2024r3dad,ye2025po3ad}, and
the current best trains PO3AD on the pseudo-anomalies of a synthesis
framework \cite{balapour2026af3ad}. Template3D-AD
\cite{liu2025template3dad} matches curvature features of the test scan
against the template at matched center points, the closest published
relative of our reference-based construction. All of them use the
benchmark's own templates, as training set or as the reference.
Zero-shot methods built on CLIP, PointAD
\cite{zhou2024pointad} and GS-CLIP \cite{deng2026gsclip}, detect
anomalies on categories they never trained on, with a few annotated
auxiliary categories as their supervision. Our learned encoders see
neither a Real3D-AD category nor a real defect.

\section{Data, Representations and Subspace Rules}
\label{sec:methods}
In this section, we describe the data, the representations in which
the transfer is studied, and the two places a subspace enters it: as a
classifier on synthetic-trained features (\cref{sec:classifier}) and
as a model of the real good parts (\cref{sec:nominalmethod}).

\subsection{Data}
\textbf{Synthetic training data.} A scan is a grid of $(x,y,z)$
samples. Following \cite{lafiosca2022}, the height is a quadric base
$z_s=\alpha x^2+\beta y^2$ plus a dent $z_d$ and noise of 0.15\,mm,
and the residual $r=z-\hat z$ from a least-squares quadric is the
input. Dents follow the parametric model of \cite{lafiosca2022} with
radius 10 to 45\,mm and depth 0.2 to 1.8\,mm. We regenerated 30{,}000
samples on a structured-light scanner's grid ($640\times760$ at
0.187\,mm per pixel), seeded by index, with the quadric fitted on all
points. IDs 0 to 23{,}999 train and the rest validate.

\textbf{Real3D-AD.} Twelve categories of small scanned objects
\cite{liu2023real3dad}: four complete good templates per category and,
after removing cut and duplicate items, 597 single-sided test scans,
302 good and 295 with a bulge or sink annotated per point. Defects
span a median of 4.5\,mm with a median peak relief of 0.6\,mm, on
objects whose own curvature dominates every window. In five categories
(seahorse, candybar, fish, toffees, car) every test scan is
pre-registered to a template within 0.05\,mm and the templates agree to
within 0.17\,mm. The other seven we register ourselves
(\cref{sec:protocol}).

\textbf{MVTec 3D-AD.} Ten classes of real structured-light scans
\cite{bergmann2022mvtec3d}, organized $800\times800$ height grids at
0.20 to 0.23\,mm pitch with per-pixel defect masks: 1{,}197 test scans,
948 defective (holes, cuts, cracks, contamination, bent parts), median
defect extent 10.7\,mm.

\subsection{Representations}
\label{sec:repr}
\label{sec:geom}
Every representation is either trained on synthetic data only or not
trained at all, and is frozen before it sees a real scan. All of them
operate on the same $64\times64$ residual windows (stride 32, boundary
band excluded).

\textbf{DentNet.} A U-Net-style FCN with a 1024-channel bottleneck at
1/16 scale, a mirrored decoder with skips, and a softmax over
\{normal, dent\}, trained with cross-entropy on the synthetic split. We
use the archived checkpoint, which reproduces its logged IoU on
independent hardware. Its probability is the end-to-end detector, and
its bottleneck, average-pooled over a window, gives the frozen feature
$\mathbf f\in\mathbb R^{1024}$.

\textbf{PointNet\texttt{++}.} A set-abstraction classifier
\cite{qi2017pointnetpp} trained from scratch on the identical windows
under the identical label rule. Each window becomes a point set
$(x,y,r)$ subsampled to 1024 points, centered and divided by its mean
radius. Two set-abstraction stages (32 centroids, 16 neighbors, widths
32-32-64, then a global stage 64-128-128) feed a two-layer head,
trained with Adam under class-balanced cross-entropy. The grouping
radius is selected on held-out \emph{training} IDs, and all three
trained encoders (radius 2.0, 0.6, 0.3) are kept as frozen
128-D representations.

\textbf{DINOv2.} A foundation-model representation that never saw a
residual image. Following \cite{lendering2026subspacead}, each window is
clipped to $\pm2$\,mm, mapped to gray and resized to 224 pixels, the
hidden states of layers $-12$ to $-18$ of DINOv2
\cite{oquab2024dinov2} are averaged, and the window's feature is the
mean of its $16\times16$ patch tokens.

\textbf{Training-free baseline.} On raw points, the surface is divided
into metric patches, each resampled to a fixed $10\times10$ lattice,
which removes point ordering and density. The smallest singular value
$S_{[2]}$ of the centered $100\times3$ coordinate matrix measures
out-of-plane variance, and the surface variation
$S_{[2]}^2/\sum_k S_{[k]}^2$ \cite{pauly2002curvature} is its
scale-invariant form. A second statistic is the signed skewness of the
plane-fit residual.

\subsection{The subspace as a classifier}
\label{sec:classifier}
\label{sec:gds}
Patch-based subspace segmentation (PBSS) classifies the frozen features
of an encoder by nearest subspace. From synthetic training windows
labeled dent above a 0.1 dent-pixel fraction, class subspaces are
built by PCA with mean $\boldsymbol\mu_c$ and basis $\mathbf V_c\in\mathbb
R^{K\times d}$, and a window is classified by nearest reconstruction,
\begin{equation}
  e_c(\mathbf f)=\bigl\|(\mathbf f-\boldsymbol\mu_c)-\mathbf V_c^\top
  \mathbf V_c(\mathbf f-\boldsymbol\mu_c)\bigr\|_2,\qquad
  \hat y=\arg\min_c e_c(\mathbf f),
  \label{eq:pbss}
\end{equation}
with signed margin $\Delta=e_{\text{normal}}-e_{\text{dent}}$. Pixel
masks come from overlap voting. Two further rules on the same features
serve as controls throughout: nearest class mean, which is
\cref{eq:pbss} with $K{=}0$, and a logistic head on standardized
features, whose margin is the logit.

\medskip\noindent\textbf{Proposition (margin stability under a feature
shift).} Let $\mathbf P_c=\mathbf V_c^\top\mathbf V_c$ be the two class
projectors. For any feature $\mathbf f$ and any shift
$\boldsymbol\delta$, each $e_c$ moves by at most
$\|(\mathbf I-\mathbf P_c)\boldsymbol\delta\|_2$, so
\begin{equation}
  \bigl|\Delta(\mathbf f+\boldsymbol\delta)-\Delta(\mathbf f)\bigr|
  \le\|(\mathbf I-\mathbf P_{\text{n}})\boldsymbol\delta\|_2
  +\|(\mathbf I-\mathbf P_{\text{d}})\boldsymbol\delta\|_2,
  \label{eq:bound}
\end{equation}
and $\Delta(\mathbf f+\boldsymbol\delta)=\Delta(\mathbf f)$ whenever
$\boldsymbol\delta$ lies in both class subspaces. The proof is the
triangle inequality, since $e_c$ is the norm of an affine function of
$\mathbf f$ whose linear part is $\mathbf I-\mathbf P_c$. The
nearest-mean margin obeys the same bound with $\mathbf P_c=\mathbf 0$,
and the logistic margin moves by exactly $\mathbf w^\top\boldsymbol\delta$,
which no subspace bounds. The margin of a nearest-subspace rule is
therefore invariant to the part of a shift inside both class subspaces
and moves by at most the part they leak.

\medskip\noindent\textbf{Leakage diagnostic.} Whether a threshold
survives depends on that move against the margin's own spread. We
therefore report the bound in units of the in-domain standard deviation
$\sigma_\Delta$ of the margin,
\begin{equation}
  L(\boldsymbol\delta)=\frac{\|(\mathbf I-\mathbf P_{\text{n}})\boldsymbol\delta\|_2
  +\|(\mathbf I-\mathbf P_{\text{d}})\boldsymbol\delta\|_2}{\sigma_\Delta},
  \label{eq:leak}
\end{equation}
and call it the leakage of the shift. With $\boldsymbol\delta$ taken as
the mean feature difference between unlabeled good target scans and the
synthetic background, $L$ needs no real defect and can be computed
before deployment. $L$ below one says the in-domain threshold moves by
less than the margin's spread on the target.

\medskip\noindent\textbf{GDS of the class subspaces.} A fourth rule
projects the features onto a generalized difference subspace
\cite{fukui2023generalized}. On unit-norm features each class subspace
$\mathbf U_c$ is built by PCA \emph{without} centering,
$\mathbf G=\sum_c\mathbf U_c^\top\mathbf U_c$, and the GDS is spanned by
the eigenvectors of $\mathbf G$ after the $R$ with the largest
eigenvalues are removed. Features are projected onto it, the class
subspaces are rebuilt there, and the margin is the difference of the
two projection energies. $K$, $R$ and the dimension $Q$ kept inside the
GDS are selected on held-out training samples only.

\subsection{The subspace as a model of the good parts}
\label{sec:nominalmethod}
When a registered reference of the good surface is available, a
subspace can model the real good parts directly. At every seed of a
lattice on the registered test scan (\cref{sec:protocol}), the
templates within tolerance give a nominal set: the template window at
the seed and at four quarter-radius in-plane jitters. The windows are
encoded by the same frozen encoder, and their $K$-dimensional PCA
subspace is the local nominal subspace, a one-class subspace
\cite{watanabe1973subspace} placed at a location. The score of a test
window is the reconstruction error of its feature in that subspace.
A score is reported raw and, where marked normalized, divided by the
median leave-one-out reconstruction error of the nominal set itself,
so that locations where the templates disagree are discounted. Two
variants change only the subspace: one is built by PCA without
centering, and the other projects the uncentered local subspaces onto
the GDS of all local subspaces of a category, built from the templates
only, with the $R$ directions of largest eigenvalue removed. The same
construction applies to the height-field patches of
\cref{sec:repr}. Nothing in a nominal set comes from a test scan.

\section{Evaluation Protocol}
\label{sec:protocol}
\textbf{Synthetic.} We report pixel IoU, window ROC-AUC and F1 on the
same validation samples, with windows labeled dent above a 0.1
dent-pixel fraction. Every interval is
a 95\% percentile bootstrap over \emph{samples}, since windows within a
sample are correlated.

\textbf{Real3D-AD.} Every choice below was fixed before any method was
scored. Two of them rest on one dataset-level statistic of the test
annotations. A probe on a 23-item subsample measured the largest
bounding-box side of the annotated defect points at a median of
5.17\,mm (4.51\,mm over all 295 defects). The primary radius, 3.6\,mm,
is 1.4 times half the probe value, never tuned on a score, and sets
the seed lattice, the
tangent frame and the localization tolerance. The similarity factor
6.6 maps the probe value onto 34\,mm, the scale of the smallest
synthetic dents, and sets the physical size of the $64\times64$ window.
No per-item annotation enters any score, threshold, nominal set or
rule. Both constants are varied: height-field statistics at 2.5, 3.6
and 5.0\,mm, the learned rules under the metric pitch, and the
reference-based stage of \cref{sec:nominal} at the annotation-free
scales 1, 2, 4 and 8. Seeds lie on a voxel lattice at stride half the
radius, on the $10\times10$ grid of \cref{sec:geom}, with a slab of
half the radius around the tangent plane applied before resampling,
since templates are two-sheeted and test scans single-sided. Object
AUROC uses the maximum over seeds and point AUROC the nearest seed's
score. Localization is the argmax seed's distance to the defect,
counted a hit within one radius. Intervals are item-clustered
bootstraps (2{,}000 item-level draws, 300 at point level). The learned
methods run zero-shot from the synthetic checkpoints on the
$64\times64$ residual window each seed defines, detrended by a
ring-fitted quadric, under the metric pitch and the similarity
transform above, with rules fitted on synthetic training features only.

\textbf{MVTec 3D-AD.} The primary protocol uses no annotation. The
grid is resampled to the synthetic scanner's pitch of 0.1875\,mm, so a
$64\times64$ window covers 12\,mm on the object as it does in training.
As the control we also resample to the pitch that maps the median
defect extent of the test masks onto 34\,mm (factor 3.19), the defect
scale, a dataset-level statistic fixed before any method was scored.
Overlapping windows with at least 95\% valid pixels are detrended by a
ring-fitted quadric, with the residual sign that gives a hole the sign of
a synthetic dent. A defect window has more than 0.1 annotated pixels, a
normal window none. Every rule's a-priori threshold is its own decision
boundary, margin zero, and the end-to-end probabilities use 0.5. The
metrics of \cref{tab:mvtec} are means over the ten classes,
with each pixel scored by the maximum over the windows covering it. The
shift $\boldsymbol\delta$ is the mean feature difference between the
windows of ten good training scans per class and the synthetic
background, and its prediction is checked on the good test scans of the
same class.

\textbf{Registration and nominal sets.} Every test scan is registered
to its category's templates by trimmed point-to-point ICP from 241
initializations per template. The median residual after registration
is 0.03 to 0.05\,mm in six categories and 0.14 to 0.24\,mm in the other
six. Six items register at 0.61 to 0.73\,mm and are kept
(\cref{sec:regsens}). Registration is the cost
of the construction, a median of 460\,s per item on 20 cores (90th
percentile 1{,}281\,s), where the encoder and the subspace scores of a
scan take under 0.1\,s each. Nominal sets are taken in the tangent
frame of each seed. Object scores use the published convention, the
maximum over every window, and a coverage rule fixed before any label
was read, the maximum over windows whose grid nodes are all sampled
(largest node-to-sample gap at most 0.25\,mm). Every constant of this stage was fixed before any
label was read and none tuned afterwards: $K{=}3$ for the local
subspace (the affine dimension of four templates), $K{=}6$ for the
height field from the declared set $\{0,1,3,6\}$, $K{=}50$ at category
level as in domain, $R{=}1$, the quarter-radius jitter, the 1.0\,mm
tolerance and the 0.25\,mm coverage gap. The sweeps of $K$, $R$ and the
template count in \cref{sec:nominal} were run afterwards.

\section{Results}
\label{sec:results}
We first confirm that every representation and rule works on synthetic
data (\cref{sec:indomain}). We then answer the two parts of the
question in turn: what a subspace contributes as a classifier on
synthetic-trained features (\cref{sec:mvtec,sec:real3d}), and what it
contributes as a model of the real good parts
(\cref{sec:nominal,sec:regsens}).

\subsection{In domain}
\label{sec:indomain}
On the validation split DentNet's dent-class IoU is 96.8\% over 4{,}835
dented samples with window AUC 1.00 and F1 0.866. On its frozen
features PBSS at $K{=}50$ reaches AUC 0.986 (F1 0.910, IoU 74.8\%),
the logistic head 0.992 (F1 0.952, IoU 76.2\%) and the nearest mean
0.954 (F1 0.714). PointNet\texttt{++} reaches 0.948 (F1 0.852, IoU
69.7\%) and the training-free $S_{[2]}$ 0.878 on the $6\times5$ grid.
Sample-clustered 95\% half-widths are at most 0.007 in AUC and 0.014 in
F1 (0.03 in AUC for $S_{[2]}$). The grouping radius
is the setting that matters most for PointNet\texttt{++} (AUC 0.89 at
radius 2.0, 0.93 at 0.6, 0.95 at 0.3). Retraining each radius with two
further initializations gives a standard deviation over the three seeds
of at most 0.001 in AUC and 0.007 in F1, so the ordering of the radii
does not come from initialization. The logistic head is the best rule
in domain, with a
paired F1 gain over PBSS of $+0.041$ \ci{0.036}{0.047}.

\subsection{The subspace as a classifier: synthetic to MVTec 3D-AD}
\label{sec:mvtec}
In this subsection, we ask what a subspace contributes when it
classifies the features of an encoder trained on synthetic data.
\Cref{tab:mvtec} applies the representations and rules of
\cref{sec:repr,sec:classifier}, fitted on the synthetic training
windows, to the 1{,}197 MVTec 3D-AD
scans.

\textbf{Ranking transfers, calibration does not.} We find that every
rule separates defect windows from normal ones on objects that share
nothing with the synthetic panels, with window AUC between 0.79 and
0.84 and pixel AUROC between 0.74 and 0.82. At the same time every rule
flags between 40 and 100\% of the good windows at its in-domain
threshold, where the end-to-end DentNet probability flags 24\%. On the
same encoder the rules differ by up to 45 points in that fraction, and
the rule that flags least changes from encoder to encoder. A
fixed-threshold comparison would therefore read \cref{tab:mvtec} as a
robustness ranking of the decision geometry that the ranking metrics
do not support.

\textbf{The mechanism.} Let $\boldsymbol\delta$ be the mean feature
difference between the windows of the good training scans and the
synthetic validation background windows. We add $\boldsymbol\delta$ to
every synthetic background feature and record how far each rule's
margin moves, in units of its in-domain standard deviation, and how
much of $\|\boldsymbol\delta\|^2$ lies inside the class subspaces. On
the DentNet encoder the shift is 15.5 feature standard deviations, of
which the normal and dent subspaces span 85 and 98\%. The leakage $L$
of \cref{eq:leak} therefore allows the margin at most 10.2 of its
in-domain standard deviations, and it moves by 4.8, where the logistic
margin, with no subspace to absorb the shift, moves by 14.9. The bound
holds on every encoder at both window scales (\cref{tab:mvtec}). The
ordering of the rules, however, depends on the shift. At the defect
scale the nearest subspace moves least of the three rules on every
task-trained encoder, whereas at the synthetic pitch, where the shift
on the two finest PointNet\texttt{++} encoders is larger and less of it
lies inside the normal subspace, it moves most. How stable a rule is
therefore depends on where the shift lies relative to its class
subspaces.

\begin{figure}[tb]
  \centering
  \includegraphics[width=0.82\linewidth]{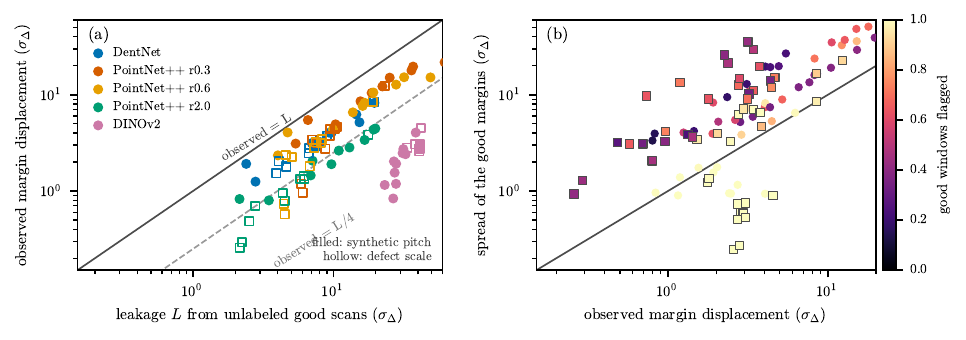}
  \caption{The leakage diagnostic on MVTec 3D-AD, one point per encoder
  and class at the synthetic pitch (filled) and the defect scale
  (hollow), nearest-subspace rule. (a) The label-free leakage $L$ of
  \cref{eq:leak} from the good training scans of a class against the
  displacement of the mean margin on its good test scans, both in
  in-domain standard deviations of the margin. (b) Displacement against
  the spread of the good test margins, colored by the flagged fraction
  of good windows.}
  \label{fig:leak}
\end{figure}

\textbf{The leakage predicts the displacement, class by class.} We
next test \cref{eq:leak} as a diagnostic (\cref{fig:leak}). For each
class, $\boldsymbol\delta$ is estimated on its ten good training scans
alone, and $L$ is compared with the displacement of the mean margin on
the good test scans of the same class, in 100 encoder, class and scale
combinations. The displacement stays below $L$ in every case, at a
median of 0.38 of it, and within each task-trained encoder $L$ ranks the classes by their displacement (Spearman 0.93 to
0.99 at both scales). On DINOv2 the bound is loose by an order of
magnitude and the ranking weaker (0.87 and 0.38). The spread of the
shifted margins matters as well. The good test features carry far more
variance than the synthetic background, so the good margins spread
over 3 to 51 in-domain standard deviations on the task-trained encoders
(\cref{fig:leak}b), and where that spread exceeds the displacement the
flagged fraction is set by both. Before deployment, the leakage gives
the displacement of the operating point and the covariance part of the
shift gives its spread.

\textbf{A different subspace construction.} The GDS rows of
\cref{tab:mvtec} test whether projecting onto the GDS of the class
subspaces (\cref{sec:gds}) absorbs more of the shift. It helps only
where the class subspaces span little of it. On the four task-trained
encoders, whose subspaces already span most of the shift, the GDS
lowers the window AUC slightly and flags more good windows. On DINOv2
features almost half of the shift lies outside the sum of the class
subspaces, the GDS discards that part, and it has the smallest margin
shift of all rules at both window scales. At the defect scale it is
also the best ranker of all rules (window AUC 0.881, image AUROC
0.621), while at the synthetic pitch the logistic head ranks 0.02
higher. The gain comes from the
projection itself: the uncentered subspace method at the same $K$
without the projection shifts by 9.37, and whitening the sum space by
$\Lambda^{-1/2}$ \cite{fukui2007komsm} instead of removing its top
lowers the window AUC to 0.75.

\textbf{Scope.} Image-level detection stays at chance for every
method, since the object's own geometry (threads, rims, edges) takes
the maximum over windows, so these numbers measure zero-shot transfer
and not detection on the benchmark's terms. The defect scale, which
uses the test masks, moves every window AUC by at most 0.08
(\cref{tab:mvtec}), and the ranking and calibration findings and the
bound hold under both scales.

\begin{table}[tb]
  \centering
  \caption{Synthetic training to MVTec 3D-AD, means over the ten
  classes, rules fitted on synthetic windows only: window AUC, pixel
  AUROC, image AUROC from the maximum over windows, good test windows
  flagged at the a-priori threshold, and the margin shift under the
  measured mean feature shift (in-domain standard deviations), at the
  synthetic pitch and at the defect scale. $K{=}50$ on DentNet and
  DINOv2, $K{=}25$ on PointNet\texttt{++}, $R$ and the GDS dimension
  $Q$ (25, 25 and 5) selected on held-out training samples.}
  \label{tab:mvtec}
  \scriptsize
  \setlength{\tabcolsep}{2pt}
  \resizebox{\linewidth}{!}{%
  \begin{tabular}{llccccrccr}
    \toprule
    & & \multicolumn{5}{c}{synthetic pitch, no annotation} & \multicolumn{3}{c}{defect scale (factor 3.19)} \\
    \cmidrule(lr){3-7}\cmidrule(lr){8-10}
    Encoder & Rule & Window & Pixel & Image & Good \% & Shift ($\sigma$) & Window & Good \% & Shift ($\sigma$) \\
    \midrule
    DentNet, end to end & probability & 0.830 & 0.807 & 0.512 & 24 & & 0.852 & 19 & \\
    \midrule
    DentNet 1024-D & nearest subspace & 0.837 & 0.808 & 0.495 & 72 & $+4.75$ & 0.852 & 98 & $+2.99$ \\
                   & nearest mean & 0.829 & 0.814 & 0.548 & 54 & $+3.08$ & 0.834 & 93 & $+3.06$ \\
                   & logistic & 0.837 & 0.808 & 0.514 & 99 & $+14.90$ & 0.853 & 100 & $+9.37$ \\
                   & GDS, $R{=}5$ & 0.823 & 0.818 & 0.534 & 77 & $+2.71$ & 0.812 & 99 & $+2.59$ \\
    \midrule
    PointNet\texttt{++} r0.3 & nearest subspace & 0.836 & 0.795 & 0.459 & 48 & $+9.79$ & 0.857 & 64 & $+2.33$ \\
                   & nearest mean & 0.830 & 0.806 & 0.561 & 54 & $+3.33$ & 0.856 & 93 & $+3.39$ \\
                   & logistic & 0.840 & 0.795 & 0.460 & 58 & $+4.93$ & 0.863 & 96 & $+3.23$ \\
                   & GDS, $R{=}1$ & 0.816 & 0.788 & 0.471 & 59 & $+3.39$ & 0.786 & 88 & $+2.82$ \\
    \midrule
    PointNet\texttt{++} r0.6 & nearest subspace & 0.838 & 0.795 & 0.460 & 46 & $+7.30$ & 0.857 & 60 & $+1.77$ \\
                   & nearest mean & 0.837 & 0.819 & 0.558 & 45 & $+3.67$ & 0.866 & 76 & $+3.53$ \\
                   & logistic & 0.839 & 0.794 & 0.471 & 49 & $+6.00$ & 0.867 & 84 & $+2.75$ \\
                   & GDS, $R{=}1$ & 0.820 & 0.788 & 0.536 & 62 & $+3.23$ & 0.812 & 84 & $+3.17$ \\
    \midrule
    PointNet\texttt{++} r2.0 & nearest subspace & 0.837 & 0.797 & 0.497 & 40 & $+1.94$ & 0.848 & 43 & $+0.56$ \\
                   & nearest mean & 0.834 & 0.795 & 0.508 & 43 & $+2.94$ & 0.867 & 64 & $+2.58$ \\
                   & logistic & 0.837 & 0.796 & 0.456 & 40 & $+1.39$ & 0.863 & 46 & $+0.93$ \\
                   & GDS, $R{=}1$ & 0.793 & 0.736 & 0.461 & 72 & $+1.90$ & 0.788 & 77 & $+1.90$ \\
    \midrule
    DINOv2 1024-D & nearest subspace & 0.814 & 0.785 & 0.479 & 100 & $+1.53$ & 0.860 & 100 & $+2.95$ \\
                  & nearest mean & 0.811 & 0.777 & 0.539 & 68 & $+1.43$ & 0.880 & 100 & $+1.67$ \\
                  & logistic & 0.825 & 0.794 & 0.513 & 100 & $+2.96$ & 0.789 & 100 & $+5.65$ \\
                  & GDS, $R{=}1$ & 0.805 & 0.774 & 0.526 & 81 & $+1.09$ & 0.881 & 100 & $+1.58$ \\
    \bottomrule
  \end{tabular}}
\end{table}

\subsection{The subspace as a classifier on Real3D-AD}
\label{sec:real3d}
On Real3D-AD the same classifiers do not transfer. The learned methods,
zero-shot from the synthetic checkpoints, stay at chance at object
level (0.50 to 0.54 at the pre-declared scale) and flag 49 to 100\% of
the windows on \emph{good} objects, and the training-free statistics $S_{[2]}$ and the
signed skew sit at 0.52 and 0.55. The defects here span a median of
4.5\,mm with a relief of 0.6\,mm on strongly curved objects, and no
statistic of a window alone separates them from the object's own
shape. The scale moves every learned method by 0.14 to 0.23 in point
AUROC, the choice of rule by 0.02. A subspace on synthetic-trained
features therefore contributes nothing here, which leads to the second
part of the question.

\subsection{The subspace as a model of the good parts: Real3D-AD}
\label{sec:nominal}
In this subsection, we ask what a subspace contributes when it models
the real good parts directly. We give the subspace methods the
reference that Real3D-AD's own baseline uses. In the five
registered categories (249 items, 124 defective) every seed has a nominal
set (\cref{sec:nominalmethod}). Feature statistics are evaluated at the
primary 3.6\,mm radius under the scale-6.6 protocol, height-field
statistics at 2.5\,mm, since the grid's interpolation error grows with
its pitch (normalized nominal subspace 0.76, 0.67 and 0.57 point AUROC
at 2.5, 3.6 and 5.0\,mm on this subset). Item-clustered half-widths
are at most 0.04 at point level and 0.07 at object level.

\textbf{Geometric statistics.} Given the reference, a CLAFIC subspace of
the nominal set's height-field patches reaches point AUROC 0.835 at
$K{=}6$ (no reference-free statistic exceeds 0.64), below the raw
point-to-template distance (0.975), since interpolation adds error of
the order of the defect.

\begin{figure}[tb]
  \centering
  \includegraphics[width=0.58\linewidth]{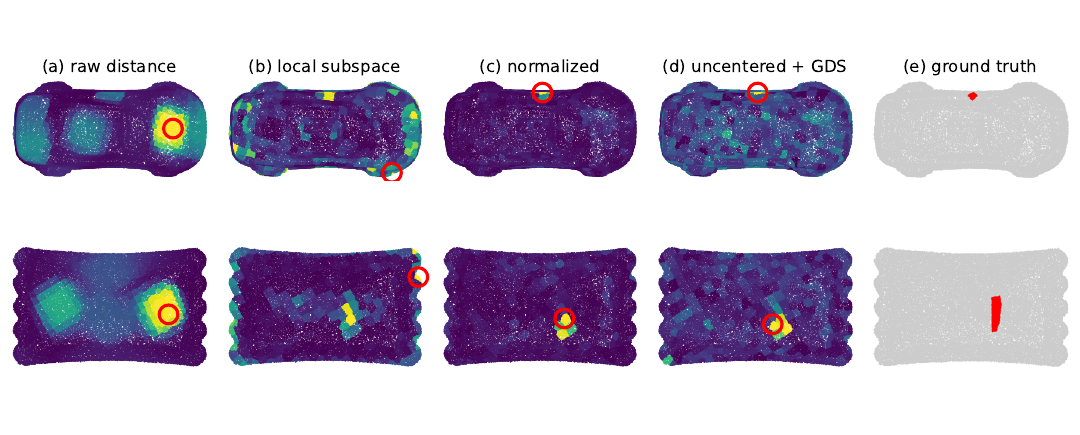}
  \caption{Two pre-registered Real3D-AD items with a sink (car, top,
  and toffees, bottom), each point colored by the score of its nearest
  seed: (a) the raw point-to-template distance, (b) the local nominal
  subspace on DentNet features, (c) the same normalized by the
  templates' leave-one-out disagreement, (d) the uncentered subspace
  projected onto the GDS ($R{=}1$) on PointNet\texttt{++} features, and
  (e) the ground truth. The red circle marks the argmax seed, 5 to
  19\,mm from the defect under (a) and (b), within 0.6\,mm under (c) and
  (d). The raw distance peaks where the scan departs from the templates
  as a whole, the unnormalized subspace on the rim where the templates
  disagree, the normalized forms on the sink.}
  \label{fig:qual}
\end{figure}

\textbf{Frozen learned features plus the reference.} The local nominal
subspace on frozen features ($K{=}3$, \cref{sec:nominalmethod})
reaches point AUROC 0.957 \ci{0.950}{0.962} on DentNet features and
0.948 on PointNet\texttt{++} features, and places the argmax within one
radius of the defect for 70 and 78\% of defective objects. The
location-specific reference produces this result: the same features
under the in-domain PBSS rule give 0.695, and a category-level
one-class subspace of all nominal windows gives 0.80. The raw
point-to-template distance ranks as
well at point level (0.975), but the local subspace localizes ten times more
sharply (a median argmax error of 0.16 against 1.6\,mm), because the
subspace absorbs the variation among the templates that a raw distance
counts as signal. At object level the aggregation rule decides: the
maximum over all windows is taken by windows that run off the object
(0.71 and 0.75), whereas the maximum over fully sampled windows gives
0.948 \ci{0.913}{0.976} and 0.910, against 0.909 for the raw distance
(\cref{fig:qual}).

\textbf{All twelve categories, on the published footing.}
\Cref{tab:compare} places our statistics beside the published Real3D-AD
results, every item under the published maximum over all windows, with
the coverage rule of \cref{sec:protocol} beside it. The published
methods train on the templates and the zero-shot methods on annotated
auxiliary categories, while ours use the templates only as the
reference, so the table compares object AUROC under one aggregation
convention and does not match training conditions. Under that convention
the local nominal subspace on DentNet features reaches a category mean
of 72.1, below the raw point-to-template distance at 76.4, and its
normalized form 78.2 \ci{74.8}{81.5}, with 78.7 on PointNet\texttt{++}
features. The gain over the raw distance therefore comes from
normalizing by the templates' own disagreement. The weak categories are
those with a registration
residual of 0.14 to 0.24\,mm, the scale of the defect relief
(\cref{sec:regsens}).

\textbf{What the subspace construction adds.} The local subspace above
is centered on its own nominal windows with the pre-declared $K{=}3$
(a sweep run afterwards peaks at $K{=}9$, so $K{=}3$ was not tuned).
Built without centering, it raises the raw category mean from 72.1 to
77.4, since the direction of the mean carries the defect amplitude that
centering removes. Projecting the uncentered local subspaces onto
their GDS and removing the one direction common to every location
($R{=}1$, fixed before scoring) gives 78.8 on DentNet features and 81.0
\ci{77.8}{84.1} on PointNet\texttt{++} features, and every $R$ from 1
to 20 stays within 1.1 points. The eight feature rows of
\cref{tab:compare} thus lie between 72.1 and 81.0 with overlapping
intervals: the choice of subspace construction moves the result by a
few points, while the reference moves it from chance to above 70.
SubspaceAD \cite{lendering2026subspacead}, one normal subspace per
category without a location, sits at 49.1 on DINOv2 features. The best
row lies above the zero-shot methods and the learned methods of the
PO3AD table, and below Template3D-AD \cite{liu2025template3dad} and
PO3AD trained on AF3AD pseudo-anomalies \cite{balapour2026af3ad}.

\textbf{Window scale without annotation.} The scale 6.6 of
\cref{tab:compare} rests on a statistic of the test annotations
(\cref{sec:protocol}). Rerunning the stage at scales fixed without
annotation, every statistic is at chance at the metric pitch, scale 1,
where a window covers 12\,mm of the object. The normalized GDS rows
reach 70.5 and 65.3 at scale 2, 79.1 and 80.3 at scale 4, and 80.9 and
80.2 at scale 8. The construction therefore needs a window within a
factor of two of the defect scale the inspection targets, which is part
of the inspection specification, and the annotation-derived scale adds
nothing over a power of two.

\begin{table}[tb]
  \centering
  \caption{Real3D-AD object-level AUROC (\%), mean over the twelve
  categories and over the five pre-registered ones. Published rows from
  \cite{ye2025po3ad} and the later papers, and two zero-shot methods
  fine-tuned on annotated auxiliary categories. Ours: every test item
  (597), no training on the
  benchmark, under the published object score (maximum over all
  windows) and the coverage rule of \cref{sec:protocol}, item-clustered
  95\% intervals in brackets.}
  \label{tab:compare}
  \scriptsize
  \setlength{\tabcolsep}{3pt}
  \begin{tabular}{lccc}
    \toprule
    Method & 12 cat., max & 12 cat., covered & 5 pre-reg., max \\
    \midrule
    BTF (raw) \cite{liu2023real3dad} & 63.5 & -- & 61.7 \\
    Reg3D-AD \cite{liu2023real3dad} & 70.4 & -- & 77.7 \\
    IMRNet \cite{li2024imrnet} & 72.5 & -- & 74.5 \\
    R3D-AD \cite{zhou2024r3dad} & 73.4 & -- & 70.4 \\
    Group3AD \cite{zhu2024group3ad} & 75.1 & -- & 83.8 \\
    PO3AD \cite{ye2025po3ad} & 76.6 & -- & 76.5 \\
    Reg2Inv \cite{yu2025reg2inv} & 78.0 & -- & -- \\
    Template3D-AD \cite{liu2025template3dad} & 84.4 & -- & -- \\
    PO3AD + AF3AD \cite{balapour2026af3ad} & 85.2 & -- & -- \\
    PointAD, zero-shot \cite{zhou2024pointad} & 74.8 & -- & -- \\
    GS-CLIP, zero-shot \cite{deng2026gsclip} & 76.4 & -- & -- \\
    \midrule
    Point-to-template distance, 2.5\,mm & 76.4 & 72.6 & 93.6 \\
    Nominal subspace, height field, norm., 2.5\,mm & 74.4 & 79.1 & 87.9 \\
    Local nominal subspace, DentNet features & 72.1 & 82.1 & 80.3 \\
    \quad normalized & 78.2 \cis{74.8}{81.5} & 80.7 & 99.0 \\
    \quad uncentered & 77.4 \cis{73.8}{80.9} & 84.6 & 92.5 \\
    \quad uncentered, GDS $R{=}1$, normalized & 78.8 \cis{75.7}{82.0} & 80.3 & 99.1 \\
    Local nominal subspace, PointNet\texttt{++} features & 74.7 & 80.9 & 89.3 \\
    \quad normalized & 78.7 \cis{75.2}{82.1} & 79.5 & 96.4 \\
    \quad uncentered & 78.3 \cis{74.8}{81.8} & 81.8 & 91.9 \\
    \quad uncentered, GDS $R{=}1$, normalized & 81.0 \cis{77.8}{84.1} & 80.7 & 97.7 \\
    \bottomrule
  \end{tabular}
\end{table}

\subsection{What the nominal subspace absorbs and what it leaks}
\label{sec:regsens}
The nominal subspace suppresses the variation its templates span, so
the remaining question is which variation of a real scan it does not
span. We answer it by measuring what the registration does to the
reference-based scores. The Spearman correlation between
a category's residual and its object AUROC is $-0.27$ for the
DentNet-feature subspace under the published convention and $-0.52$
under the coverage rule, and pooled item AUROC falls from 0.76 to 0.34
from the residual bin below 0.05\,mm to the one above 0.3\,mm. On the
five pre-registered categories we perturb the registered pose by a
translation (0.1 to 1.0\,mm) and a rotation (0.25 to 2 degrees) about
the centroid and rescore against this set's own baseline of 0.952 point
and 0.80 object AUROC. The largest translation takes point AUROC to
0.77 and object AUROC to 0.64, the largest rotation to 0.90 and 0.68.
Point AUROC first falls below 0.90 at a translation of
0.5\,mm, the scale of the relief. A translation of 0.1\,mm, well inside
the in-plane jitter the nominal set contains, costs 0.003. One at the
scale of the relief is the variation the subspace does not span.

\section{Discussion}
\label{sec:discussion}
\textbf{Findings.} We asked what subspace representations contribute
to synthetic-to-real transfer, and the answer depends on where the
subspace enters. As a classifier on synthetic-trained features, a
subspace ranks real defects no better than a nearest mean or a
logistic head, and every rule ranks them at window AUC 0.79 to 0.84 on
MVTec 3D-AD. Its contribution is a margin that is less sensitive to the
part of the shift inside its class subspaces, and the leakage outside
them, measured on unlabeled good scans, bounds and ranks how far the
threshold moves (\cref{eq:leak}). The shift measured here is too large
for any threshold to survive, and its covariance part spreads the good
margins further. On frozen DINOv2 features, where the class subspaces
span under half of the shift, a GDS of them is the least sensitive
rule. As a model of the real good parts, a subspace of registered
templates turns a frozen encoder into a detector on Real3D-AD, where
the classifier fails. There the reference and its registration
contribute most, the normalization by template disagreement next, and
the choice of subspace construction a few points.

\textbf{Limitations.} For the subspace as a classifier, we measured the
shift of one benchmark on five representations of one training task,
and its class-conditional part, the shift of real defect features, is
not measured. One foundation model was tested, on a representation it
was not trained for. For the subspace as a model of the good parts,
MVTec 3D-AD has no registered references, so the construction was
tested on Real3D-AD only. There the seed radius and the primary window
scale are dataset-level statistics of the test annotations, fixed
before scoring. Annotation-free scales reproduce \cref{tab:compare}
within 1.5 points, but the radius was not swept. The result is bounded
by registration, which is also its computational cost.

\textbf{Implications for subspace methods.} A subspace is invariant
only to the variation it spans, so its value under synthetic-to-real
shift can be read from the geometry of the shift before deployment.
For a classifier, the leakage and the spread of the shifted margins,
both measurable from unlabeled good scans, decide whether its threshold
survives. For a model of the good parts, the variation it spans is the
disagreement among templates and the jitter of the nominal set, and the
variation it leaks is a pose error at the scale of the defect relief,
which is why the registration bounds the result. On Real3D-AD the
oldest construction, a few good examples at the same location compared
by reconstruction error on a frozen representation, carried most of the
result, and the uncentered and GDS variants moved it by a few points.
One-class subspaces of normal data have long served as anomaly
detectors with fixed, untrained features, on ECG signals
\cite{ye2012flac} and histopathological images \cite{nosato2011hlac},
and our result places the same construction at a location on 3-D
surfaces.

\textbf{Open directions.} For the classifier, the leakage and the
spread of \cref{fig:leak} can recalibrate the threshold, or choose $R$
for the GDS, before a real defect is labeled, and comparing a window
and its neighbors to the class subspaces by canonical angles, the
setting of the mutual subspace method, would replace a single feature
with a set. For the model of the good parts, a seed radius fixed from
the templates alone and a registered reference for MVTec 3D-AD would
remove the remaining use of test annotations and test the construction
on a second benchmark. A kernel GDS, convex cones for the sign every
reconstruction score discards, a second 2-D foundation model and a
point-cloud pretrained encoder would extend both parts.

\section{Conclusion}
\label{sec:conclusion}
We asked what subspace representations contribute to synthetic-to-real
transfer, using 1{,}197 MVTec 3D-AD scans and 597 Real3D-AD objects. A
subspace contributes in two places and in different ways. As a
classifier on synthetic-trained features it ranks real defects no
better than other rules, and its contribution is a threshold whose
displacement is bounded by the leakage of the shift outside its class
subspaces, a quantity measurable before deployment from unlabeled good
scans. As a model of the real good parts it makes a frozen encoder
detect real defects, with point AUROC 0.957 on the pre-registered scans
and a category-mean object AUROC of 78 to 81 on every item, with no
training on the benchmark. In that setting the reference and its
registration set what the detector can reach.

\end{document}